\documentclass[graybox]{styles/svmult}

\usepackage{amssymb}
\usepackage{amsmath}
\usepackage[justification=centering]{caption}
\usepackage{multirow}
\usepackage{array}
\usepackage{adjustbox}
\usepackage{xcolor}
\usepackage{colortbl}
\usepackage{pifont}
\newcommand{\cmark}{\ding{51}}%
\newcommand{\xmark}{\ding{55}}
\usepackage{hyperref}
\usepackage[normalem]{ulem}

\usepackage{mathptmx}       
\usepackage{helvet}         
\usepackage{courier}        
\usepackage{type1cm}        
\usepackage{makeidx}         
\usepackage{graphicx}        
\usepackage{multicol}        
\usepackage[bottom]{footmisc}

\makeindex             

\begin{document}

\title*{Visually Guided UGV for Autonomous Mobile Manipulation in Dynamic and Unstructured GPS Denied Environments}
\titlerunning{Autonomous Mobile Manipulation} 
\author{Mohit Vohra and Laxmidhar Behera}
\institute{Indian Institute of Technology, Kanpur, \{mvohra, lbehera\}@itk.ac.in}

\maketitle

\abstract*{A robotic solution for the unmanned ground vehicles (UGVs) to execute the highly complex task of object manipulation in an autonomous mode is presented. This paper primarily focuses on developing an autonomous robotic system capable of assembling elementary blocks to build the large 3D structures in GPS-denied environments. The key contributions of this system paper are \textit{i)} Designing of a deep learning-based unified multi-task visual perception system for object detection, part-detection, instance segmentation, and tracking, \textit{ii)} an electromagnetic gripper design for robust grasping, and \textit{iii)} system integration in which multiple system components are integrated to develop an optimized software stack. The entire mechatronic and algorithmic design of UGV for the above application is detailed in this work. The performance and efficacy of the overall system are reported through several rigorous experiments.}

\abstract{A robotic solution for the unmanned ground vehicles (UGVs) to execute the highly complex task of object manipulation in an autonomous mode is presented. This paper primarily focuses on developing an autonomous robotic system capable of assembling elementary blocks to build the large 3D structures in GPS-denied environments. The key contributions of this system paper are \textit{i)} Designing of a deep learning-based unified multi-task visual perception system for object detection, part-detection, instance segmentation, and tracking, \textit{ii)} an electromagnetic gripper design for robust grasping, and \textit{iii)} system integration in which multiple system components are integrated to develop an optimized software stack. The entire mechatronic and algorithmic design of UGV for the above application is detailed in this work. The performance and efficacy of the overall system are reported through several rigorous experiments.}
\section{Introduction}
Robotics applications are designed to have a transformational impact on day-to-day life in areas including disaster management, healthcare, household works, transportation, construction, and manufacturing \cite{vohra2019real} \cite{pharswan2019domain}. International robotics challenges such as the Amazon Robotics Challenge (ARC) and Mohamed Bin Zayed international robotics challenge (MBZIRC) are setting new benchmarks to advance the state-of-the-art in autonomous solutions for these robotics applications. In addition to the development of new algorithms, system integration is an essential step to complete the task. Therefore, in this work, we will describe our system architecture for the task of assembling large 3D structures in GPS-deprived environments. However, the proposed system architecture for UGV-based object manipulation is not limited to the above settings and can be deployed for other real-world applications.

\begin{figure}
\sidecaption[t]
\centering
  \includegraphics[width=4.5cm, height=3cm]{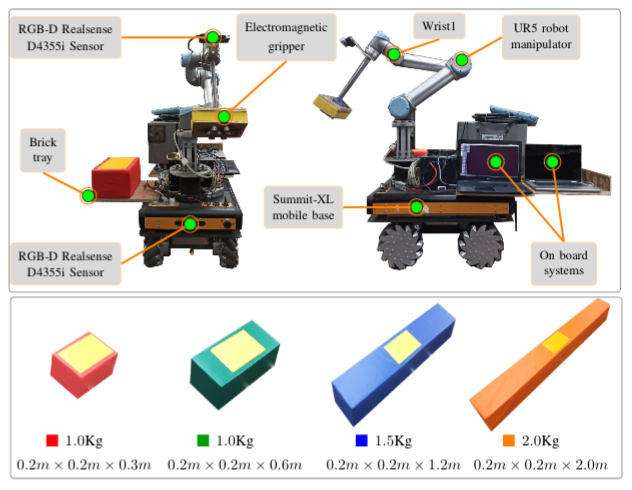}
  \caption{Top: UGV endowed with arm and gripper. Bottom: Bricks used for the task.}
  \label{fig:UGV}
\end{figure}

For the task of assembling a structure, a UGV needs to search, track, grasp, transport and assemble the bricks according to a user-specific pattern. Apart from that, for a GPS-denied environment, the system should harness the visual information for localization and should also be endowed with a suitable grasping mechanism. While performing the task, a mobile manipulator may need to move across several positions in a large workspace. It encourages us to use the onboard computational devices for various algorithms. Further, the top of each brick consists of a ferromagnetic region ($0.15m\times0.25 m$) shaded with yellow color called grasping region and is attached in order to facilitate electromagnetic grasping as shown in Fig.\ref{fig:UGV}.
\par
In order to perform the task, the perception system must locate the grasping region on bricks. To equip the UGV with the functionality of brick searching, tracking, grasping and placing, in a limited computational resources, we present a deep learning based multi-task visual perception system. The state-of-the-art object detectors \cite{yolov1} \cite{SSD} predicts a bounding box for an object. Since, the bounding box may contain significant non-object regions, therefore, an additional computation step is required to extract the object information, hence, we prefer detection (or tracking) by segmentation. Since, we already have the segmentation information for each object, therefore, to save computations, we prefer segmentation based object tracking. \cite{wang2019fast}, can track the object by predicting the object mask at each instant, but it relies on bounding box initialization. The key advantages of the tracking by segmentation are \textit{i)} The output of the brick detection by segmentation network can be directly used for tracker initialization, \textit{ii)}The output of the tracking network can be directly used for tracking in next subsequent frames, and  \textit{iii)} The detection network and tracking network can be integrated into a single network, as both networks are performing a segmentation task. Apart from the vision system, we also have designed a self-adjusting electromagnetic gripper, which, when pressed against the brick surface, can adjust itself without loading the end-effector motors. It is done to align  the magnetic gripper appropriately with the grasping regions to ensure a firm grasp. Overall, the main contributions of this work can be summarized as below:

\begin{enumerate}
    \item Multi-task deep neural network based visual perception system for brick detection, part detection, instance segmentation and tracking.
    \item Development of an self-adjusting electromagnetic gripper.
    \item System integration and a state machine to realize a fully autonomous UGV based robotic manipulation system.
\end{enumerate}

\section{Related Work} \label{sec:related_work}


\textbf{Object detection:}
RCNN is one of the first successful object detectors. The subsequent versions of RCNN are Fast-RCNN and Faster-RCNN, which have less inference time as compared to RCNN. In addition RetinaNet \cite{lin2017focal}, FCOS \cite{tian2019fcos} perform detection by predicting the axis-aligned bounding boxes. In general, the predicted box may contain significant background. Hence, to reduce the background impact in the detection, various solutions have been presented in the literature. For example, in \cite{li2018multiscale}, the author predicts a rotated bounding box to maximize the object regions inside the predictions. Similarly, Mask-RCNN \cite{maskrcnn} predicts the mask of target object inside the bounding box predictions.

\textbf{Object Tracking:}
Discriminant correlation filter (DCF) based trackers \cite{henriques2014high} localize the object by computing the correlation between the filter and the search window. Siamese neural network based trackers \cite{bertinetto2016fully} localizes the object by calculating the similarity between the target image and the search window. Some deep trackers directly regress the bounding box parameters of the object in the next frame by comparing the target object patch and the search window \cite{held2016learning}. The third type of tracker uses a classifier based approach to localize the object in the next frame. In this, a classifier assigns a score to each target candidate, which is generated at different scales and random locations around the previous object location \cite{nam2016learning}, and a target candidate with maximum score is considered as the new object location.


\textbf{Object Part Detection:}
Recent approaches \cite{shi2020points} have demonstrated part detection in point cloud data for 3D object detection. Conversely, \cite{lorenz2019unsupervised} has detected object parts in the image plane for the task of detection, tracking, and segmentation. Since in the task, we have to detect yellow areas (ferromagnetic regions) from the brick surface, and we have to classify whether it is related to green brick, blue brick, or red brick. Therefore, it is intuitive to detect objects (or bricks) in the image plane and then detect yellow regions corresponding to each brick region, which leads us to integrate the function of object detection and part detection in a single network.
\section{System Description} \label{sec:mechtronics}

\textbf{UGV Platform:}
The UGV hardware system used for the task is shown in Fig. \ref{fig:UGV}. The hardware setup consists of a UR$5$ robotic arm deployed on a ROBOTNIK Summit-XL mobile base. The UR$5$ arm is a $6$-DoF industrial arm having an end-effector position accuracy of $\pm 0.1mm$ and a payload capacity of $5$kg. The mobile base has four high power motor mecanum wheels with a maximum payload of $100$Kg. The mecanum wheel allows the base to move in any direction while keeping the front in a constant direction. The UGV is equipped with two RGB-D sensors, One sensor is mounted on the $wrist1$ of the UR$5$ arm, and the other sensor is mounted at the base's front face, shown in Fig. \ref{fig:UGV}. The raw images captured by the first sensor is used for extracting the crucial information about the bricks while the data from the second sensor is used for the UGV localization.

\begin{figure}
\sidecaption[t]
\centering
  \includegraphics[width=6cm, height=1.5cm]{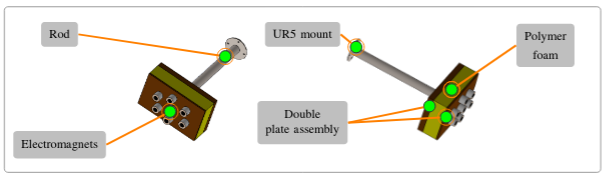}
 \caption{Proprietary electromagnetic gripper.}
  \label{fig:gripper}
\end{figure}

\textbf{Gripper Design:}
Fig. \ref{fig:gripper} shows our carefully designed electromagnetic gripper for grasping the desired objects. It consists of a double plate assembly with a polymer form in between them. On one plate, we have attached seven electromagnets with uniform distribution at the center of the gripper.  All the electromagnets are connected to the battery through a switch. The switch can be controlled by transmitting signals high or low on the ROS topic. The double plate assembly is connected to the UR5 end-effector through a steel-rod. The presence of the foam in the gripper allows it to adjust according to the tilted brick surface when pressed against the brick surface. Hence we call the gripper as a self-adjusting gripper. Further, to avoid any damage to the gripper, we keep measuring the force exerted on the gripper. If the value exceeds some threshold, then grasping operation is complete; otherwise, we keep pushing the gripper against the brick surface.
\section{Multi-Task Unified Visual Perception block} \label{sec:visual}

\begin{figure}
\centering
  \includegraphics[width=9cm, height=2.5cm]{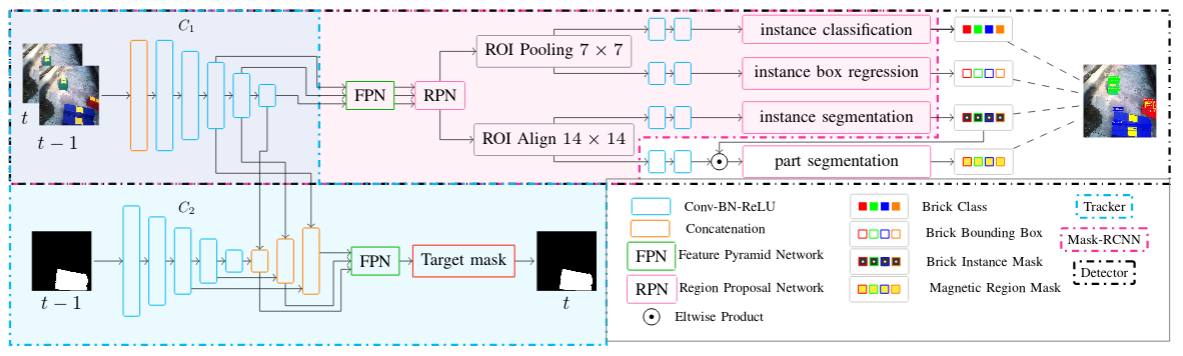}
 \caption{Visual Perception Pipeline }
   \label{fig_cnn}
\end{figure}

The main task of the visual perception block is brick detection, part segmentation, instant segmentation and brick tracking. Limited computational resources prohibit us to use multiple CNN for the above. Hence, we aim to unify the above four tasks in a deep neural network based visual perception framework. Fig. \ref{fig_cnn} represents the proposed perception block and in following section, we briefly discuss each of the component individually.

\textbf{Instance Detection and Segmentation:} Since there are multiple instances of the bricks, hence it is necessary to perform instance detection and segmentation. The top most choice for instance level segmentation and detection is Mask-RCNN\cite{maskrcnn}. Being very memory exhaustive, we modified the architecture for our computational constrain system. First, we carefully design a CNN backbone similar to AlexNet, which is represented by $C_1$. Further, with $3$-$4$ iterations of weight pruning, we ended up with a light-model which can infer in run-time. In the baseline model of Mask-RCNN, for object detection the features upto the stage-$2$ of ResNet are used. In our setup, the object detection features are limited to stage-$3$.  Further, the remaining of the architecture for task of instance detection and for segmentation remains the same.

\textbf{Part Detection:}
Since we have designed a lightweight MaskRCNN for object detection and segmentation. Further, we have a requirement of object part localization, i.e, ferromagnetic regions. Hence we aim to extend the previously designed CNN for the above task. In order to achieve this, features from ROI-Align layer are passed through the convolution layers as shown in Fig. \ref{fig_cnn}. The features from convolutional layers are multiplied with the output of the instance segmentation branch.  Finally, a binary mask is learned using binary cross entropy loss. For more detailed explanation, the reader is advised to refer the original work of Mask-RCNN.

\textbf{Conditional Tracking:}
After the detection of all target brick instances, we aim to localize the bricks in the subsequent frames by predicting the binary mask corresponding to the target bricks. At this instant a binary mask $M_{t-1}$ corresponding to the target brick in the image $I_{t-1}$ is generated and fed to the unified network $C_2$. In the next step, the current image $I_{t}$ is fed to the network, and based on previous image $I_{t-1}$, and previous binary mask $M_{t-1}$, new mask $M_{t}$ is predicted. Since the new prediction is guided by previous mask, hence we call the mask as support mask.
\par
An isolated AlexNet style CNN ($C_2$) (similar to the object detector) is designed to extract spatial feature embedding corresponding to the support mask as shown in Fig. \ref{fig_cnn}. To meet the memory constraints, the architecture $C_2$ has very less number of parameters in comparison with architecture $C_1$. Further, both tracker and detector shares the CNN backbone $C_1$ for extracting the high dimensional embedding from RGB images.
\par
Now, the features from stage-$3$, stage-$4$ and stage-$5$ of both $C_1$ (RGB images) and $C_2$ (support mask) are merged by executing a concatenation operation. Concatenation operation essentially guides the tracker to predict in next frame by incorporating features from RGB image. Now, these merged features are passed through FPN block for final mask predictions.

\section{The Complete Workflow} 
\label{sec:path_planning}

At the start of the task, the UGV is provided with a target pattern for assembling the bricks at the assembly area. By parsing the target pattern file, UGV extracts the target brick ID and performs certain set of operations to bring the target brick from the piles to the assembly area. The complete robotic motion can be divided into various modules, depending on the state of the UGV, which are explained below.

\begin{figure}
\sidecaption[t]
\centering
  \includegraphics[width=6cm, height=2.8cm]{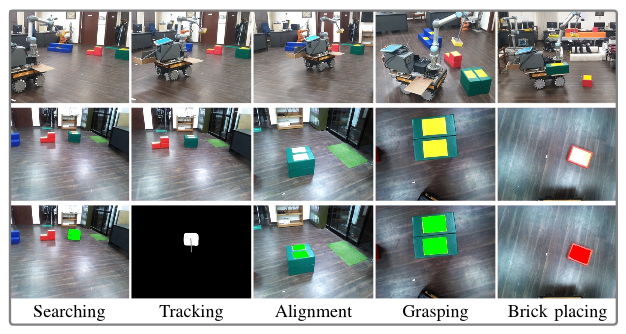}
 \caption{UGV state at different modules, Row1: view from external camera, \\Row2: UR5 arm sensor image, Row3: Network output}
 \label{fig:complete pipeline}
\end{figure}

\textbf{Searching Module:} In this state, current image is fed to the perception system, which performs bricks detection and segmentation (Fig. \ref{fig:complete pipeline}, Col. 1). If the network ensures the presence of the target brick, then the search module is complete, otherwise, the UGV system will explore the space by rotating the base by $45^{\circ}$ in a clockwise direction and again execute the search module.
    
\textbf{Tracking Module:} In tracking module, perception system predicts the binary mask $M_t$ corresponding to the target brick ID in the current image $I_t$. Further, to ensure that UGV is moving in right direction, the lateral velocity of the UGV is controlled by the horizontal-component of the displacement vector formed between the target brick centroid and the image frame centroid as shown in  Fig. \ref{fig:complete pipeline}, Col. 2. Further, the forward velocity of the UGV is proportional to the distance of the target bricks from the UGV and it gradually decreases to zeros when distance is less than threshold distance ($l_{th}$).
    
\begin{figure}
\centering
  \includegraphics[width=8cm, height=2cm]{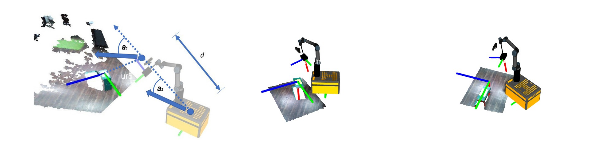}
 \caption{Alignment and brick placing operation}
  \label{fig:alignment_module}
\end{figure}

\textbf{Alignment Module:} In this module, the UGV will align its position wrt the brick as the UR5 arm has a span of $1.0m$ only. As a first step, the current image is fed the perception system for ferromagnetic regions segmentation (Fig. \ref{fig:complete pipeline}, Col. 3). By applying PCA (Principal Component Axis) on the point cloud data corresponding to ferromagnetic regions, the major and minor axis (green and blue axis in Fig. \ref{fig:alignment_module}, Col. 1) are estimated. The PCA-axis along with the $3$D centroid of the point cloud data represents the $6$D pose of the brick. Based on the current pose of UGV and the $6$D brick pose, piecewise motion planner is executed which consists of three steps; \textit{i)} Rotate base by ${a_2}$, \textit{ii)} Move forward by distance $d$, and \textit{iii)} Rotate base by ${a_1}$,  where, ${a_2}$, ${d}$, ${a_1}$ are estimated online. The final positioning of UGV after alignment is shown in Fig. \ref{fig:alignment_module}, Col. 2.
    
\textbf{Grasping Module:} Firstly, $6$D pose of the brick is estimated (Fig. \ref{fig:complete pipeline}, Col. 4). Due to the sensor noise, the estimated pose is not accurate, hence to compensate for this, the gripper is pressed against the brick surface till the force exerted on the gripper reached a value ($f_{th}$). The foam present in the gripper allows it to adjust according to the tilted brick surface. Thus force feedback based grasping, with foam in the gripper, allows us to compensate for the sensor noise.
    
\begin{figure}
\sidecaption[t]
\centering
  \includegraphics[width=6cm, height=5.5cm]{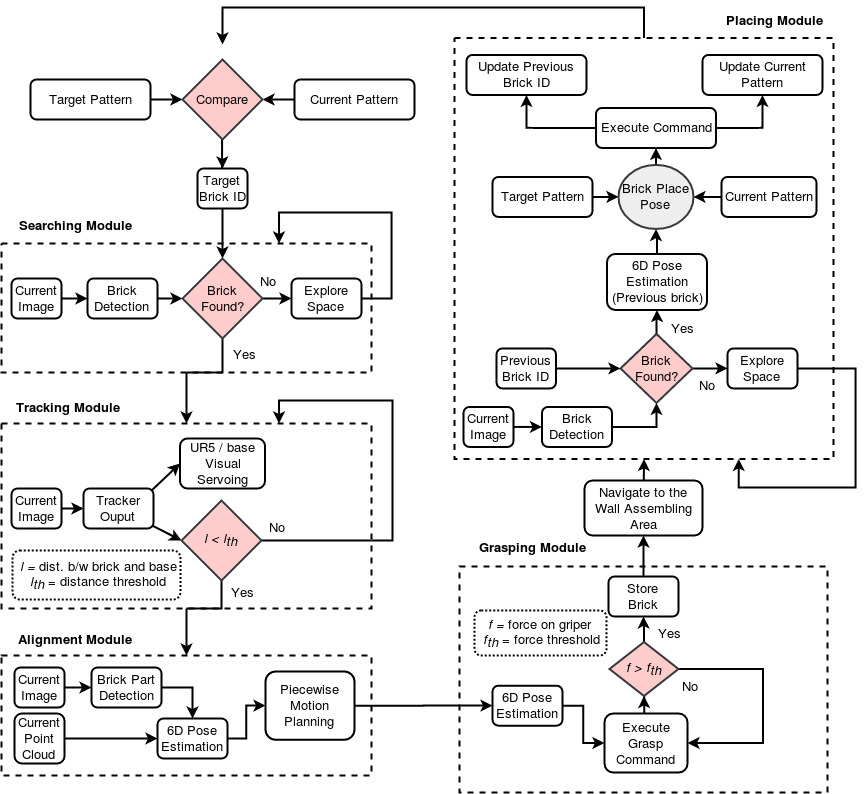}
 \caption{Simplified State Machine of the overall System}
  \label{fig:Flow}
\end{figure}

\textbf{Placing Module:} After the grasping, UGV navigate to the wall assembly area. For navigation, we have used the visual-slam module by incorporating the state-of-the art real-time ORB-SLAM$2$\cite{orb2} on UGV. The final place pose for the grasped brick depends on the previously placed brick. Hence, $6$D pose of the previous brick is calculated. Based on the previous brick pose, target pattern, and current pattern or current brick ID, the place pose is calculated. After the brick placing operation, the current pattern is updated. Fig. \ref{fig:complete pipeline}, Col. 5 represents the UGV state in the placing module. Here the system has aligned wrt to the previous brick (red brick) and again computing the $6D$ pose of the (red) brick for estimating the new pose for placing the green brick. The simplified flow of the overall system is shown in Fig. \ref{fig:Flow}.

\section{Experiments} \label{sec:experiment}
To demonstrate the validity and usefulness of the work presented in this paper, we conducted several set of experiments to evaluate the performance of the proposed visual perception network. Further, the evaluation of the overall system performance is also provided, where we have performed a series of experiments that provide a statistical analysis of the system's performance.

\textbf{Dataset:}
In this section we will explain the procedure for generating the dataset for two tasks namely detection and tracking.
\begin{itemize}
    \item \textbf{Detection} The collection of data procedure is based on the work \cite{deepquick}. Following the idea of \cite{deepquick}, a small sized dataset of about $100$ images is collected, where each image has multiple brick instances. The dataset was split between training set and testing set in the proportion of $7:3$. Now, for each brick instance, the box regression and the segmentation is performed, while for ferromagnetic regions, only the segmentation is required. During the data generation process, the mask corresponding to the brick instance and for ferromagnetic regions are marked manually. These masks are used as ground truths for instance segmentation and brick-part segmentation. The ground truths for box regression is generated by fitting a rectangle on the manually marked object instance mask. In order to augment the dataset, synthetic data augmentation technique is used heavily \cite{deepquick}.
    
    \item \textbf{Tracking} For tracking, $10$ indoor video sequences as well as $10$ outdoor sequences, each at the rate of $\sim30$FPS are collected. The video sequences has the brick instances whose size varies as we move closer to the bricks. Further, the video sequences are down-sampled to the rate of $\sim2$ FPS and each frame is annotated for instance detection, classification, segmentation and part segmentation. The complete dataset for tracking contains $200$ images, where $150$ images are used for training and $50$ are used for testing. Further, the dataset has been augmented using the synthetic augmentation technique \cite{deepquick}.

\end{itemize}

\textbf{Training Policy:}
We consider the architecture of baseline network as Arch$_1$ and we also consider two other variants i.e. Arch$_2$ and Arch$_3$ which contains $25\%$ and $50\%$ of additional number of parameters as compared to Arch$_1$. The kernel sizes corresponding to each layers are mentioned in Table-\ref{tab_kernelsize}. Since object detection and tracking are two different tasks, hence separate or individual training procedure is required. In order to avoid the individual training, brick instance masks and brick-part masks are also annotated with the object masks in the training dataset. In this way, the detector can be trained using the tracking dataset. Further, the training images are chosen with equal probability from both dataset i.e., the detector dataset and tracker dataset. With this technique, the deep neural network can get the experiences of both temporal data (because of video sequences) and the single image object detection. Further, the following hyperparameters are used for training, that is, base learning rate is set to  $0.001$,  we have used the learning rate policy=$step$, $ADAM$ optimizer is used with parameters $\beta_1=0.9$ and $\beta_2=0.99$.

\begin{table}
\centering
\caption{Kernel sizes. `$*$' corresponds to input channels}
\label{tab_kernelsize}

\tiny

\arrayrulecolor{white!60!black}

\begin{tabular}{c|c|c}
\hline
Layer & $C_1$ & $C_2$ \\ \hline
Stage-$1$ & $4\times3\times3\times3$ & $2\times1\times3\times3$  \\
Stage-$2$ & $8\times4\times3\times3$ & $4\times2\times3\times3$ \\
Stage-$3$ & $16\times8\times3\times3$ & $4\times4\times3\times3$  \\
Stage-$4$ & $32\times16\times3\times3$ & $8\times4\times3\times3$  \\
Stage-$5$ & $32\times32\times3\times3$ & $16\times8\times3\times3$  \\
Others & $12\times*\times3\times3$ & $4\times*\times3\times3$  \\ \hline
\end{tabular}
\end{table}

\textbf{Ablation Study:}
Table-\ref{tab_performance} shows the performance of the deep neural network for various tasks. The accuracy of network along with the timing performance for half-Precision (FP$16$) and for single-precision (FP$32$) computations on a single GeForce GTX $1050$, $4$GB GPU, is reported. Further, the performance of all the three architectures Arch$_1$, Arch$_2$, and Arch$_3$ are reported. From the Table-\ref{tab_performance}, we observed that Arch$_2$ shows an minor improvement over Arch$_1$, similarly, Arch$_3$ demonstrates slight improvements over Arch$_1$, despite the $50\%$ more parameters. This is because, the basic network architecture for all variants are same except for the number of parameters. And another reason is that, the object used for experiments have significant differences, and thus intra-class variance is extremely low.

\begin{table}
\caption{ Performance Analysis}
\label{tab_performance}
\centering

\tiny

\begin{tabular}{c|c|c|c|c|c}
\hline

Network  & Time & Box & Seg & Part Seg &  Tracker \\ 
Architecture  & (ms) & mIoU & mIoU & mIoU &  mIoU \\ 
\hline
Arch$_1$ (FP$16$) & \textcolor{blue}{$49$} & $80.3$  & $79.2$  & $73.1$ & $83.4$ \\ 
Arch$_1$ (FP$32$) & $123$ & $82.2$  & $80.0$  & $72.9$ & $84.8$ \\ \hline
Arch$_2$ (FP$16$) & $80$ & $81.2$  & $79.7.$ & $73.6$ & $84.1$ \\ 
Arch$_2$ (FP$32$) & $161$ & $83.5$  & $81.1$  & $74.0$ & $85.6$ \\ \hline
Arch$_3$ (FP$16$) & $101$ & $85.3$  & $80.6$  & $73.5$ & $84.4$ \\ 
Arch$_3$ (FP$32$) & $187$ & \textcolor{blue}{$86.7$}  & \textcolor{blue}{$81.9$}  & \textcolor{blue}{$74.1$} & \textcolor{blue}{$86.4$} \\ \hline
\end{tabular}  
\end{table}

\textbf{Synthetic Data Augmentation:}
Table-\ref{tab_augmentation} demonstrates the impact of synthetic scenes and other augmentations as described in \cite{deepquick}. From Table-\ref{tab_augmentation}, we observed that by incorporating the synthetic augmentation technique, we can see the significant improvement in the performance. Further, the tracker performance degradation is observed when the blurring effect is included in the augmentation process.

\begin{table}

\caption{ Effect of Synthetic Data Augmentation for Arch$_1$, FP$16$}
\label{tab_augmentation}

\setlength\tabcolsep{3pt}
\centering

\tiny

\begin{tabular}{c c c c c c|c c c c}

\hline
\multicolumn{6}{c|}{Data Augmentation} & \multicolumn{4}{|c}{mIoU score} \\ \hline

Color & Scale & Mirror &  Blur & Rotate & Synthetic scene & Detection & Segmentation & Part Segmentation & Tracker \\ \hline

 \xmark  & \xmark & \xmark & \xmark & \xmark & \xmark & \multicolumn{1}{|c}{$23.7$} & $31.8$ & $18.8$ & $15.3$\\ 
 \cmark  &  &  &  &  & & \multicolumn{1}{|c}{$25.3$} & $32.1$ & $21.3$ & $17.7$ \\ 
 \cmark  & \cmark &  &  &  &  &\multicolumn{1}{|c}{$30.1$} & $38.6$& $29.1$& $23.1$\\ 
 \cmark  & \cmark & \cmark &  &  &  &\multicolumn{1}{|c}{$32.3$} & $40.2$& $31.9$ & $25.5$\\ 
 \cmark  & \cmark & \cmark & \cmark &  &  &\multicolumn{1}{|c}{$32.5$} & $41.4$& $33.8$ & \textcolor{red}{$24.1$}\\ 
 \cmark  & \cmark & \cmark & \cmark & \cmark &  &\multicolumn{1}{|c}{$37.2$} & $49.8$ & $37.7$ & $28.4$\\ 
 \cmark  & \cmark & \cmark & \cmark & \cmark & \cmark & \multicolumn{1}{|c}{\textcolor{blue}{$80.3$}} & \textcolor{blue}{$79.2$} & \textcolor{blue}{$73.1$} & \textcolor{blue}{$83.4$}\\ \hline
 
\end{tabular}  
\end{table}

\textbf{Unified vs Non-Unified Perception System:}
The main contribution of this work is the unification of various task. Hence, it is very intuitive to compare the performance of multi-task network against the individual network for same tasks namely detection, part segmentation and tracking. The individual network architectures have the same configurations as that of multi-task network. It can be observed from the Table-\ref{tab_unified} that the unified pipeline is way better than the individual networks in terms of timing performance. Further the memory consumed by the unified network is $40$\% of total memory consumed by all three networks. The link for one of the experiment is \url{https://youtu.be/nPiMzrBFJ-A}.

\begin{table}
\caption{Unified Vs Non-Unified Perception System, Arch$_1$, FP$16$}
\label{tab_unified}
\centering

\tiny

\begin{tabular}{c|c|c|c|c}
\hline

Network  & \multicolumn{4}{c}{Time(ms)} \\ \cline{2-5}
Architecture  & \multicolumn{1}{c|}{detection} & \multicolumn{1}{c|}{Part Seg} & \multicolumn{1}{c|}{Tracking} & \multicolumn{1}{c}{Total} \\ 
\hline
Arch$_1$ &  $--$  & $--$  & $--$ & \textcolor{blue}{$49$} \\ \hline
Detection & $26$  & $--$  & $--$ & \multirow{2}{*}{$106$} \\ 
Part Segmentation & $--$  & $35$  & $--$ & \\
Tracking & $--$ & $--$  & $45$ & \\ \hline
\end{tabular}  
\end{table}

\textbf{Overall System Evaluation:}
The complete robotic system was tested for $50$ rounds. In each round, the pile's initial position is at different locations and has different configurations, and the robotic system has to assemble the 3D structure at the assembly area according to the target pattern. In each round, the target pattern consists of $6-8$ bricks. Thus overall the UGV is tested for $346$ number of bricks and for each brick, we count the number of times the individual module has performed satisfactory. The performance of the module is defined as unsatisfactory if the final pose of the UGV, after the execution of the specific module is random. On the other hand, if the module performs satisfactory, then UGV will land at appropriate pose and can easily switch to next module. Table-\ref{sys_per} summarize the performance of UGV in our experiments.

\begin{table}
\centering
\caption{Overall System Evaluation}
\label{sys_per}

\tiny

\begin{tabular}{c|c}
\hline
\textbf{Module} & Success Trials \\ \hline
Searching & 346/346 (\textbf{100.0\%}) \\ \hline
Tracking & 346/346 (\textbf{100\%})\\ \hline
Alignment & 292/346 (\textbf{84.4\%})\\ \hline
Grasping & 292/292 (\textbf{100.0\%}), 292/346 (\textbf{84.4\%}) \\ \hline
Placing & 218/292 (\textbf{74.6\%}), 218/346 (\textbf{63.0\%}) \\ \hline
Overall & 218/346 (\textbf{63.0\%}) \\ \hline
\end{tabular}
\end{table}




From Table-\ref{sys_per}, we observed that the searching module and tracking module has $100\%$ accuracy. This is because the dataset is small and network has been trained with extensive data augmentation technique. The alignment module is successful for $292$ times out of $346$ trials, which shows $84\%$ success rate. This is because, the front sensor gets exposed to the textureless view, and thus visual-slam fails to update the pose of the UGV. For the remaining successful trials, grasping module was tested and it has shown the $100\%$ success rate and overall the system score is $292/346$ which is $84\%$. Further the placing module is successful in $218/292$, as during the brick placing operation, visual slam fails to update the pose of UGV. The overall system score is $218/346$. From the experiments, we observed that one of the main point which can enhance the performance of UGV is to estimate the accurate state of the UGV during alignment module and brick placing module. For this, in future we aim to localize the UGV by using the data from multiple sensors, which can be integrated at different body parts of the UGV.
\section{Conclusion} \label{sec:conclusion}
We have presented a robotic solution to enable unmanned ground vehicles (UGVs) to perform the highly complex task of assembling the elementary blocks (or bricks) in the GPS-denied environment. The proposed system consists of a deep learning based unified multi-task visual perception system for the tasks of instance detection, segmentation, part segmentation and object tracking. The perception system can infer at the rate of $20$ FPS on a single GeForce GTX $1050$ $4$GB GPU. The proposed visual perception module has been extensively tested for various test cases which includes indoor and outdoor environments and the performance of the perception module for various tasks is reported in this paper. The propose perception module is integrated onto the UGV system for the task of brick assembly. Further to facilitate the grasping of bricks, an electromagnetic based self-adjusting gripper is designed, and a force based feedback grasping strategy is deployed to compensate for the sensor noise.  Further, the complete robotic system is tested for $50$ rounds of bricks assembly task, and the proposed system has shown the accuracy of $63\%$. The primary reason for the low accuracy is that visual-slam fails to update the state of the UGV when the depth sensor get exposed to the textureless view. Hence by incorporating the multiple sensors for localization, accuracy can be increased, which will be considered as future work.
\bibliography{citations} 

\begin{thebibliography}{10}

\bibitem{vohra2019real}
M.~Vohra, R.~Prakash, and L.~Behera, ``Real-time grasp pose estimation for
  novel objects in densely cluttered environment,'' in {\em 2019 28th IEEE
  International Conference on Robot and Human Interactive Communication
  (RO-MAN)}, pp.~1--6, IEEE, 2019.

\bibitem{pharswan2019domain}
S.~V. Pharswan, M.~Vohra, A.~Kumar, and L.~Behera, ``Domain-independent
  unsupervised detection of grasp regions to grasp novel objects,'' in {\em
  2019 IEEE/RSJ International Conference on Intelligent Robots and Systems
  (IROS)}, pp.~640--645, IEEE, 2019.

\bibitem{yolov1}
J.~Redmon, S.~Divvala, R.~Girshick, and A.~Farhadi, ``You only look once:
  Unified, real-time object detection,'' in {\em Proceedings of the IEEE
  conference on computer vision and pattern recognition}, pp.~779--788, 2016.

\bibitem{SSD}
W.~Liu, D.~Anguelov, D.~Erhan, C.~Szegedy, S.~Reed, C.-Y. Fu, and A.~C. Berg,
  ``Ssd: Single shot multibox detector,'' in {\em European conference on
  computer vision}, pp.~21--37, Springer, 2016.

\bibitem{wang2019fast}
Q.~Wang, L.~Zhang, L.~Bertinetto, W.~Hu, and P.~H. Torr, ``Fast online object
  tracking and segmentation: A unifying approach,'' in {\em Proceedings of the
  IEEE conference on computer vision and pattern recognition}, pp.~1328--1338,
  2019.

\bibitem{lin2017focal}
T.-Y. Lin, P.~Goyal, R.~Girshick, K.~He, and P.~Doll{\'a}r, ``Focal loss for
  dense object detection,'' in {\em Proceedings of the IEEE international
  conference on computer vision}, pp.~2980--2988, 2017.

\bibitem{tian2019fcos}
Z.~Tian, C.~Shen, H.~Chen, and T.~He, ``Fcos: Fully convolutional one-stage
  object detection,'' in {\em Proceedings of the IEEE international conference
  on computer vision}, pp.~9627--9636, 2019.

\bibitem{li2018multiscale}
S.~Li, Z.~Zhang, B.~Li, and C.~Li, ``Multiscale rotated bounding box-based deep
  learning method for detecting ship targets in remote sensing images,'' {\em
  Sensors}, vol.~18, no.~8, p.~2702, 2018.

\bibitem{maskrcnn}
K.~He, G.~Gkioxari, P.~Doll{\'a}r, and R.~Girshick, ``Mask r-cnn,'' {\em CVPR},
  2017.

\bibitem{henriques2014high}
J.~F. Henriques, R.~Caseiro, P.~Martins, and J.~Batista, ``High-speed tracking
  with kernelized correlation filters,'' {\em IEEE transactions on pattern
  analysis and machine intelligence}, vol.~37, no.~3, pp.~583--596, 2014.

\bibitem{bertinetto2016fully}
L.~Bertinetto, J.~Valmadre, J.~F. Henriques, A.~Vedaldi, and P.~H. Torr,
  ``Fully-convolutional siamese networks for object tracking,'' in {\em
  European conference on computer vision}, pp.~850--865, Springer, 2016.

\bibitem{held2016learning}
D.~Held, S.~Thrun, and S.~Savarese, ``Learning to track at 100 fps with deep
  regression networks,'' in {\em European Conference on Computer Vision},
  pp.~749--765, Springer, 2016.

\bibitem{nam2016learning}
H.~Nam and B.~Han, ``Learning multi-domain convolutional neural networks for
  visual tracking,'' in {\em Proceedings of the IEEE conference on computer
  vision and pattern recognition}, pp.~4293--4302, 2016.

\bibitem{shi2020points}
S.~Shi, Z.~Wang, J.~Shi, X.~Wang, and H.~Li, ``From points to parts: 3d object
  detection from point cloud with part-aware and part-aggregation network,''
  {\em IEEE Transactions on Pattern Analysis and Machine Intelligence}, 2020.

\bibitem{lorenz2019unsupervised}
D.~Lorenz, L.~Bereska, T.~Milbich, and B.~Ommer, ``Unsupervised part-based
  disentangling of object shape and appearance,'' in {\em Proceedings of the
  IEEE Conference on Computer Vision and Pattern Recognition},
  pp.~10955--10964, 2019.

\bibitem{orb2}
R.~Mur-Artal and J.~D. Tard{\'o}s, ``Orb-slam2: An open-source slam system for
  monocular, stereo, and rgb-d cameras,'' {\em IEEE Transactions on Robotics},
  vol.~33, no.~5, pp.~1255--1262, 2017.

\bibitem{deepquick}
A.~Kumar and L.~Behera, ``Semi supervised deep quick instance detection and
  segmentation,'' in {\em 2019 International Conference on Robotics and
  Automation (ICRA)}, pp.~8325--8331, IEEE, 2019.

\end{thebibliography}
\bibliographystyle{ieeetr}
\end{document}